\documentclass[10pt,twocolumn,letterpaper]{article}

\usepackage{3dv}
\usepackage{times}
\usepackage{epsfig}
\usepackage{graphicx}
\usepackage{amsmath}
\usepackage{amssymb}

\usepackage{amsthm}
\usepackage{bm}
\usepackage{amsfonts}
\usepackage{mathtools}
\usepackage{subfig}
\usepackage{booktabs}
\usepackage{verbatim}
\usepackage{makecell}

\usepackage[pagebackref=true,breaklinks=true,letterpaper=true,colorlinks,bookmarks=false]{hyperref}

\threedvfinalcopy 

\def\threedvPaperID{230} 

\ifthreedvfinal\fi
\begin{document}

\title{Localising In Complex Scenes Using Balanced Adversarial Adaptation}

\author{Gil Avraham\thanks{Authors contributed equally} 
\and
Yan Zuo$^{*}$ 
\and
Tom Drummond \\
ARC Centre of Excellence for Robotic Vision, Monash University, Australia \\
{\tt\small \{gil.avraham, yan.zuo, tom.drummond\}@monash.edu}}

\maketitle
\thispagestyle{empty}

\begin{abstract}
Domain adaptation and generative modelling have collectively mitigated the expensive nature of data collection and labelling by leveraging the rich abundance of accurate, labelled data in simulation environments. In this work, we study the performance gap that exists between representations optimised for localisation on simulation environments and the application of such representations in a real-world setting. Our method exploits the shared geometric similarities between simulation and real-world environments whilst maintaining invariance towards visual discrepancies. This is achieved by optimising a representation extractor to project both simulated and real representations into a shared representation space. Our method uses a symmetrical adversarial approach which encourages the representation extractor to conceal the domain that features are extracted from and simultaneously preserves robust attributes between source and target domains that are beneficial for localisation. We evaluate our method by adapting representations optimised for indoor Habitat simulated environments (Matterport3D and Replica) to a real-world indoor environment (Active Vision Dataset), showing that it compares favourably against \emph{fully-supervised} approaches.
\end{abstract}


\begin{figure}[t]
\centering
\includegraphics[width=0.45\textwidth]{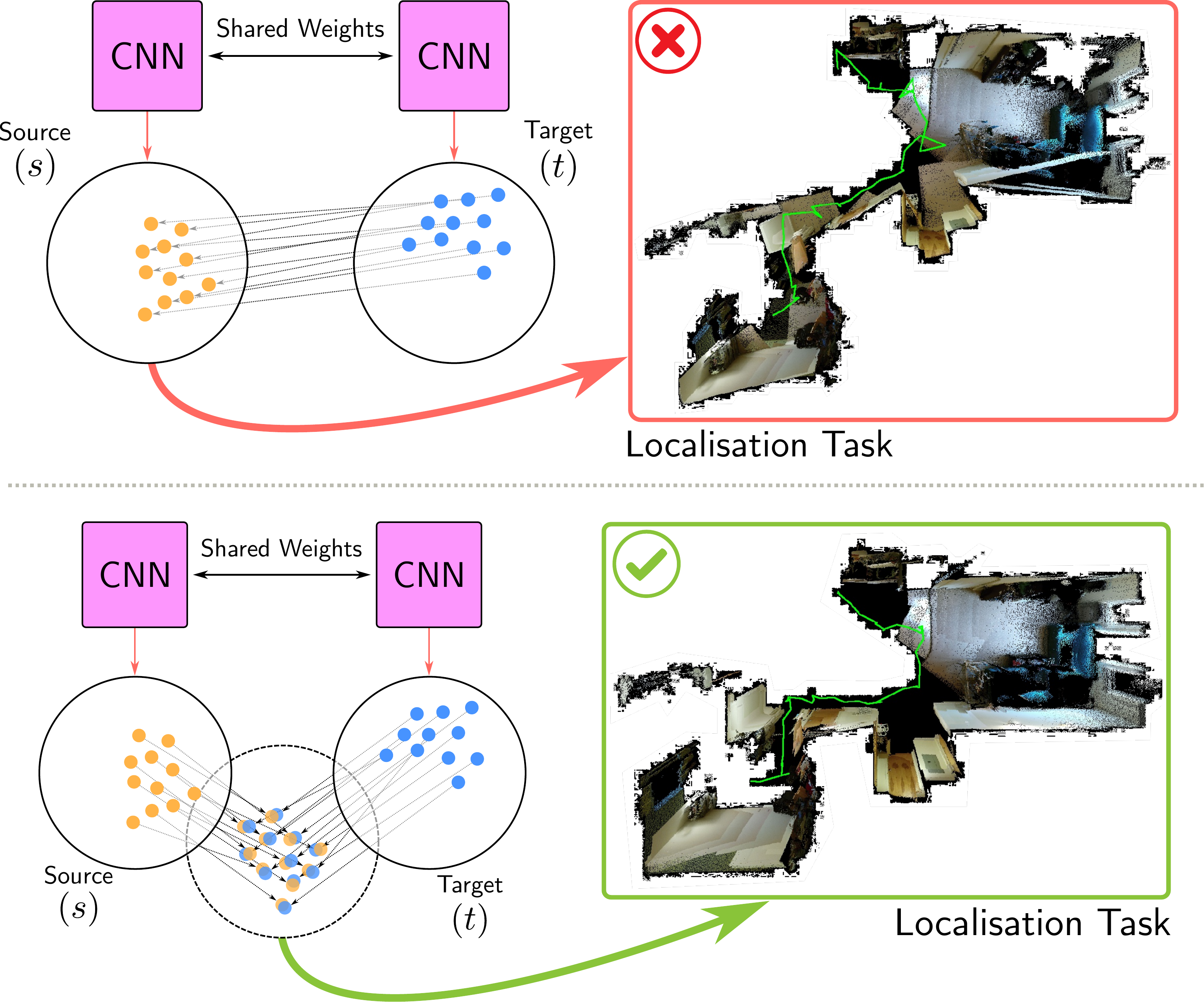}
\caption{The Balanced Adversarial Adaptation (BAA) Network learns a mapping from a source and target domain to a shared domain. 
BAA bridges the generalisation gap between domains by leveraging cheap and accurate representations acquired from a simulation environment and adapting these representations to operate on real-world environments.}
\label{fig:frontpage}
\end{figure}

\section{Introduction}
Despite deep learning largely entrenching itself as a cornerstone of modern computer vision, the technique commonly relies on training models which require heavy supervision with labelled data that is costly to obtain~\cite{liu2017survey}. Whilst simulation offers potential refuge from this problem, this data often does not closely resemble real-world data. Subsequently, when models trained on simulated data are deployed in the real-world, large performance gaps are often observed~\cite{wang2018deep}. Domain adaptation aims to bridge this gap, although modern approaches are largely restricted to image understanding tasks such as recognition, segmentation and detection~\cite{he2016deep,long2015fully,redmon2016you}, where it has found success across these tasks through training on a source domain and leveraging data from the target domain to minimise the performance gap between domains. This domain gap in image understanding extends into the area of inference in 3D space and geometry of a scene, where tasks such as monocular visual odometry (VO) has seen a renewal of interest with the advent of deep representational learning~\cite{wang2017deepvo,clark2017vinet}.

Performing this task well is crucial towards creating fully autonomous agents, where spatial awareness about the scene plays a key role in assisting with applications such as navigation and human-robot interaction~\cite{thrun2004toward,goodrich2008human}. This is an active area of research where supervised learning of the localisation task has resulted in frameworks that learn robust representations which carry the geometric properties of the scene~\cite{bloesch2018codeslam,zhou2018deeptam,clark2017vidloc}. Although these approaches have demonstrated promising results, they require copious amounts of training data which are both difficult to obtain and biases the model towards becoming highly attuned towards a specific domain. Despite this, it is apparent that the rich representations offered by deep learning are desirable for building effective systems for tackling difficult visual odometry tasks (such as large camera baseline shifts between frames)~\cite{henriques2018mapnet,avraham2019empnet}. 

In this paper, we explore a novel method to adapt between a simulated and real-world domain for the task of localisation. In particular, our framework adapts directly on the feature representations used for localising using a symmetrical adversarial technique, where both the source and target domain representations are mapped to a shared ``in-between" domain (see Fig.~\ref{fig:frontpage}).
The representation extractor is encouraged to optimise the representations that both conceal the domain from which they came from and also preserve mutual geometrical information between the domains which are used to estimate an agent's pose. We show the effectiveness of this approach, where we use our Balanced Adversarial Adaptation (BAA) Network to adapt from simulation environments to real-world environments and show localisation performance that is on-par or surpasses \textit{supervised models trained on real-world data}.
Our approach offers the following contributions:
\begin{itemize}
\item A novel, bi-directional adaptation approach which utilises a symmetrical adversarial technique. This symmetrical approach yields demonstrably improved adapted localisation over methods which use uni-directional adaptation. 
\item Empirically, we show that our weakly-supervised model (trained on simulation data) can perform on-par or even outperform \textit{supervised} methods for localising on the challenging real-world Active Vision dataset.
\item To the best of our knowledge, this is the first work to explore domain adaptation in the context of localisation. Our work establishes the baseline for any future research on this important task.
\end{itemize}



\section{Related Work}
\label{sec:related_work}
\subsection{Localisation}
The task of localisation involves estimating the camera pose from observed image data. Initial frame-by-frame methods such as~\cite{costante2015exploring,kendall2015posenet} treated the localisation task as a regression problem and directly optimised on the camera pose, where the former included optical flow between pairs of sequential frames as additional information to the input. Later works exploited the geometric properties of the scene, using these as constraints to restrict the search space for optimisation of their models~\cite{mahjourian2018unsupervised,dharmasiri2018eng}.~\cite{bloesch2018codeslam} used latent codes to capture scene geometry, using an encoder-decoder setup to jointly optimise for both depth and camera pose.

Related sequence-to-sequence methods maintain an internal memory of previous observations~\cite{wang2017deepvo,clark2017vinet} where a Convolution Neural Network (CNN) and an LSTM~\cite{hochreiter1997long} is used for feature extraction and inference of dynamics across sequences of frames respectively.~\cite{zhou2018deeptam} proposed to optimise for depth estimation and camera pose together, through regressing on the camera pose and estimating a cost volume for reconstructing the environment.~\cite{clark2017vidloc} uses bi-directional LSTMs to smoothly estimate camera pose and localise over video sequences.

Map-based approaches attempt to explicitly incorporate previously visited states internally where a fixed sized latent map from a 2D top-down viewpoint perspective is maintained~\cite{zhang2017neural,parisotto2017neural}.~\cite{gupta2017cognitive} extended up on this approach towards exploring real-world environments, modifying the global map representation to an egocentric one. MapNet~\cite{henriques2018mapnet} further extended upon the approach of~\cite{gupta2017cognitive}, which maintained a ground-projected allocentric map. The approach used a brute-force localisation approach where the current state was localised against each patch of the maintained allocentric map and in doing so, temporal information was lost in this process, where non-viable areas of the map were considered for localising.~\cite{avraham2019empnet} addressed this issue through the maintaining a short term memory point-embedding map to localise against incoming states. Most recently,~\cite{li2020self} proposed an online meta-learning algorithm which enabled a VO framework to continuously adapt to new environments in a self-supervised manner.

\subsection{Domain Adaptation}
A popular type of transfer learning is the task of domain adaption. Domain adaptation methods aim to address the detrimental effects of domain shift through learning a model which reduces this shift between the source and target distributions~\cite{pan2009survey}. Shallow adaptation methods used fixed feature representations, building a mapping between the source and target domains~\cite{daume2009frustratingly,bruzzone2009domain,chu2013selective,gong2013connecting,pan2010domain,gheisari2015unsupervised,pachori2018hashing}. These shallow approaches were also used in conjunction with deep neural networks to extract representations which were used to align feature sub-spaces between domains~\cite{hoffman2013one,raj2015subspace,nguyen2015dash,zhang2017deep}

Recently, Deep Domain Adaptation has emerged as a method of choice when dealing with large amounts of unlabelled data. Deep domain adaptation was first proposed by~\cite{ganin2014unsupervised}, where both domain adaptation and feature learning were combined within a single training process. This enabled the learning of representations that were invariant to changes in domain. Most modern approaches to domain adaptation learn transformations that map features from the source domain to the target domain's feature space~\cite{tzeng2015simultaneous,long2015learning,bousmalis2016domain}. 

Commonly, modern deep domain adaptation methods utilise adversarial training as part of their process~\cite{tzeng2017adversarial,peng2018zero,liu2016coupled,tzeng2015adapting}. In particular, Generative Adversarial Networks (GANs)~\cite{goodfellow2014generative} are often directly incorporated into adversarial mapping approaches and several training setups encourage the learning of domain-invariant features in this manner~\cite{zhu2017unpaired,taigman2016unsupervised,reed2016generative,zhang2017stackgan,hoffman2017cycada}.

\section{Preliminaries}
\label{sec:background}
As this work addresses the problem of adapting representations that are suited towards localisation, the choice of representations is critical for this task. For an RGB image $\mathcal{I} \in \mathbb{R}^{H \times W \times 3}$, it is desirable to have a mapping function $\mathcal{F}: \mathcal{I} \rightarrow \mathcal{H}$ that can transform the given image $\mathcal{I}$ to spatial representations $\mathcal{H} \in \mathbb{R}^{H_r \times W_r \times C}$, where $H_r$ and $W_r$ denote the respective height and width of a representation tensor and $C$ denotes the number of channels. Furthermore, if we observe every spatial point in $\mathcal{H}$, $\ell_{i} \in \mathbb{R}^{C}$ as a list of vectors $\{\ell_{i}\}_{i=0}^{H_{r}W_{r}}$, then a single vector representation is sampled from: $\ell_{i} \sim p_{\mathcal{F}}(\ell|P_{i})$, where $P_i \in \mathbb{R}^{H_{u} \times W_{u} \times 3}$ corresponds to a patch in the image $\mathcal{I}$. For the task of localisation, ideally the mapping function $\mathcal{F}$ should be optimised such that:
\begin{enumerate}
\item For a spatial representation $\mathcal{H}$, two representation vectors are mutually exclusive (\textit{i.e.} $\{\forall \ell_{i},\ell_{j} \in \mathcal{H} \land i \neq j, 0=\textless\ell_{i},\ell_{j}\textgreater\}$). This ensures the uniqueness of a local descriptor that allows representations across a sequence of images to be matched distinctively~\cite{ke2004pca}.
\item Given a representation vector $\ell_{1,i}$ which corresponds to a patch $P_{1,i}$ in image $I_{1}$, applying an homography on an image $\mathcal{I}_{2} = H\mathcal{I}_{1}$, yields a corresponding patch $P_{2,j}$ with representation vector $\ell_{2,j}$. If the patches $P_{1,i}$ and $P_{2,j}$ correspond to the same 3D location, then the mapping function $\mathcal{F}$ should yield $\ell_{1,i} = \ell_{2,j}$~\cite{szeliski2010computer}.
\end{enumerate}

\subsection{A Primer on EMP-Net}
\label{ssec:pose_invaraint_representations}
EMP-Net~\cite{avraham2019empnet} provides a recent baseline which meets the requirements covered above. The mapping function $\mathcal{F}$ is optimised such that every representation vector $\ell_{i}$, corresponds to a 3D location in a scene, and the representations are mutually exclusive. This is achieved by first constructing a ground-truth correspondence matrix $\mathcal{C}_{gt,t}$ at time $t$, between a sequence of frames. The operation is performed by taking every frame $F_{t}$ in a sequence, which contains RGB-D information and the pose, $T_{t} \in \mathbb{SE}(3)$, and using camera intrinsics $K$, depth information $\mathcal{D}_{t}$ and pose $T_{t}$ for generating 3D point correspondences relative to previous frames $\{F_{t'}\}_{t'=t-b}^{t-1}$ (a buffer of size $b=4$ is used in \cite{avraham2019empnet}). This formulation produces a sparse correspondence matrix $\mathcal{C}_{gt,t}$, where every 3D point in frame $F_{t}$ corresponds only to matching 3D points in frames $\{F_{t'}\}_{t'=t-b}^{t-1}$. 

By using this correspondence matrix $\mathcal{C}_{gt,t}$, as one-hot ground-truth labels, a mapping $\mathcal{F}$ can be optimised to produce representations that yield an inferred correspondence matrix $\mathcal{C}_{f,t}$. This is performed by computing a spatial representation map $\mathcal{H}_{t-b} = \mathcal{F}(F_{t-b})$, where $\mathcal{F}$ is implemented using a convolutional neural network. A list of vector representations $\{\ell_{i,t-b}\}_{i=0}^{H_{r}W_{r}}$ is obtained from the spatial map $\mathcal{H}_{t-b}$. By iteratively performing this operation up to $F_{t}$, the inferred correspondence matrix $\mathcal{C}_{f,t}$ is produced by taking the negative pairwise distance between $\{\ell_{i,t}\}_{i=0}^{H_{r}W_{r}}$ and $\{\{\ell_{i,t'}\}_{i=0}^{H_{r}W_{r}}\}_{t'=t-b}^{t-1}$ and by applying a Softmax operation, we obtain a probability distribution for every representation vector $\ell_{i,t} \in \mathcal{H}_{t}$. For optimising the mapping function $\mathcal{F}$, a cross-entropy loss is taken between the ground-truth correspondence matrix and the inferred correspondence matrix as follows:
\begin{equation}
CE_{loss} = -\frac{1}{H_{r}W_{r}}\sum_{j=1}^{H_{r}W_{r}}\sum_{i=1}^{bH_{r}W_{r}}\mathcal{C}_{gt,t}[i,j]\log\mathcal{C}_{f,t}[i,j]
\label{eq:emp_loss}
\end{equation}

\begin{figure*}[t]
\centering
\includegraphics[width=0.95\textwidth]{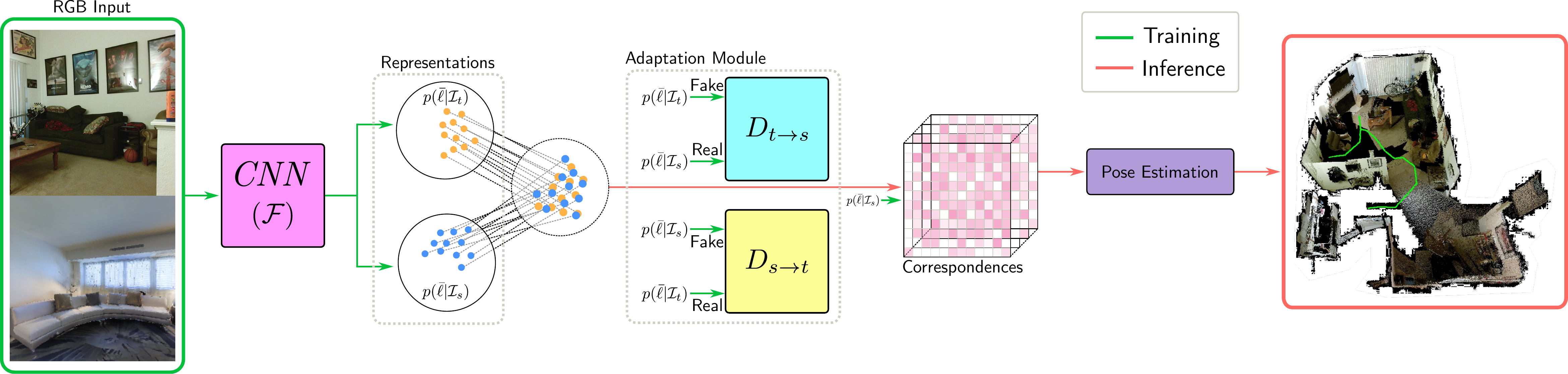}
\caption{The BAA system is optimised by learning representations which are effective for localising in a source domain and simultaneously applies a symmetrical adversarial loss on both source and target domains. The convolutional neural network ($\mathcal{F}$) employs EMP-Net~\cite{avraham2019empnet} for extracting representations used for pose estimation, while concurrently acting as a generator in an adversarial setup. BAA's adaptation module uses a dual discriminator setup, $D_{s \rightarrow t}$ and $D_{t \rightarrow s}$, which have complementing objectives. The representations are fed in a swapped order to each of the discriminators. This setup forces the generator ($\mathcal{F}$) to conceal each of the domain representations into a shared domain. In the training phase, source domain information is available as RGB-D trajectories with ground-truth poses for learning representations that are effective for pose estimation. The adaptation module receives RGB images from both source domain and target domain and pushes both the learned source representations $p(\bar{\ell}|\mathcal{I}_{s})$ and target representations $p(\bar{\ell}|\mathcal{I}_{t})$ to a shared domain, which is invariant to both domains. At inference stage, given a trajectory of a RGB sequence from the target domain, the representations from the shared domain are used to perform pose estimation.}
\label{fig:system_overview}
\end{figure*}

\section{Balanced Adversarial Adaptation}
\label{sec:dual_adversarial_adaptation}

\subsection{Problem Statement}
\label{ssec:problem_statement}
In this adaptation problem, there exists source and target domains. The information available in the source domain contains $N_{s}$ sequences of length $L$. Each sequence captures RGB-D frames $\{I_{s,i},D_{s,i}\}_{i=1}^{L}$, and pose information $\{T_{s,i}\}_{i=1}^{L}$. The source domain is typically a simulated environment where the measurements are noiseless sensory information. The information available in the target domain is a list of $N_{t}$ RGB images $\{I_{t,i}\}_{i=1}^{N_{t}}$, without any specific order. The target domain is typically a real-world environment, and the available data are RGB camera images taken from different real-world scenes (not sequences). Using the method outlined in Section~\ref{ssec:pose_invaraint_representations}, we can optimise a mapping function $\mathcal{F}$ to obtain pose invariant representations for the source domain which have desirable characteristics for estimating the camera pose. For convenience, we denote the concatenation of these representations as $\bar{\ell} \sim p_{\mathcal{F}}(\bar{\ell}| \mathcal{I}_{s})$, where $\mathcal{I}_{s}$ denotes a random variable of an RGB image from the source domain.

Alternatively, we can view our source and target data through a careful assumption that they share a hidden variable $\mathcal{L}$ which describes a property that underlies both domains (\textit{i.e.} $I_{s} \sim p_{s}(\mathcal{I}_{s} | \mathcal{L})$ and $I_{t} \sim p_{t}(\mathcal{I}_{t} | \mathcal{L})$. For the case of indoor scenes, $\mathcal{L}$ could represent the geometry (shape information) of a scene upon which the texture is conditioned, and either a simulated or real image can be sampled.

Without performing any adaptation, the representations $\bar{\ell} \sim p_{\mathcal{F}}(\bar{\ell}| \mathcal{I}_{t})$ will not be effective for localising on the target domain, as they were optimised to operate on the source domain. Our main objective is to obtain representations for the target domain so that they can be used for localisation in a real-world setting. The goal is to find a procedure that re-optimises the mapping function $\mathcal{F}$ so that when provided with an image $\mathcal{I}_{t}$ from the target domain, the extracted representations will exhibit the desired attributes outlined in Section~\ref{sec:background}.

\subsection{Conditional Dual Adversarial Adaptation}


The mapping function $\mathcal{F}$ is initially optimised to produce representations for source domain data (\textit{i.e.} $\mathcal{I}_{s})$. Our objective is to re-optimise our mapping function $\mathcal{F}$ which provides representations invariant to the domain of the input. 
Given both source and target distributions:
\begin{equation}
\begin{split}
I_{s} \sim p_{s}(\mathcal{I}_{s} | \mathcal{L}) , I_{t} \sim p_{t}(\mathcal{I}_{t} | \mathcal{L})
\label{eq:data_dists}
\end{split}
\end{equation}
and a mapping function which emits the two distributions:
\begin{equation}
\begin{split}
\bar{\ell}_{s} \sim p_{\mathcal{F}}(\bar{\ell}| \mathcal{I}_{s}), \bar{\ell}_{t} \sim p_{\mathcal{F}}(\bar{\ell}| \mathcal{I}_{t})
\label{eq:representations_dist}
\end{split}
\end{equation}
we use the dual adversarial objective proposed by~\cite{nguyen2017dual}:
\begin{equation}
\begin{split}
\min_{\mathcal{F}}\max_{D_{1},D_{2}} V(\mathcal{F},D_{1},D_{2}) = \alpha E_{\bar{\ell} \sim p_{\bar{\ell}|\mathcal{I}_{s}}}[\log D_{1}(\bar{\ell})] \\ 
+ E_{\bar{\ell} \sim p_{\bar{\ell}|\mathcal{I}_{t}}}[-D_{1}(\bar{\ell})] \\
+ E_{\bar{\ell} \sim p_{\bar{\ell}|\mathcal{I}_{s}}}[-D_{2}(\bar{\ell})] \\
+ \beta E_{\bar{\ell} \sim p_{\bar{\ell}|\mathcal{I}_{t}}}[\log D_{2}(\bar{\ell})]
\label{eq:dual_optimisation}
\end{split}
\end{equation}
which is equivalent to optimising our mapping function $\mathcal{F}$ when both discriminators are trained to convergence, $D_{1}^{*}, D_{2}^{*}$:
\begin{equation}
\begin{split}
\min_{\mathcal{F}} V(\mathcal{F},D_{1}^{*},D_{2}^{*}) = \alpha(\log \alpha - 1) + \beta(\log \beta - 1) \\
\alpha D_{KL}(p(\bar{\ell}|\mathcal{I}_{s})||p(\bar{\ell}|\mathcal{I}_{t})) + \beta D_{KL}(p(\bar{\ell}|\mathcal{I}_{t})||p(\bar{\ell}|\mathcal{I}_{s}))
\label{eq:dual_optimisation_KL}
\end{split}
\end{equation}
where we have two $KL$-divergence terms which allow us to ``push'' the distributions in both directions and the applied force can be controlled using the $\alpha, \beta$ terms.
The mapping function will minimise the objective $\min_{\mathcal{F}}V \iff p(\bar{\ell}|\mathcal{I}_{s}) = p(\bar{\ell}|\mathcal{I}_{t})$, so that $\bar{\ell}$ is invariant to the domain in which it was originated from.

Given that the model uses an additional regularisation term for the mapping function $\mathcal{F}$ which optimises the representations for localisation in the source domain data (see Section~\ref{ssec:dual_adversarial_architecture}) where we have labelled information, we can use $\alpha$ and $\beta$ to dictate which $KL$-divergence is activated in the optimisation process. This encourages the mapping function $\mathcal{F}$ to conceal the non-mutual information between domains whilst retaining their mutual information (\textit{i.e.} the hidden variable $\mathcal{L}$ useful for the task). For localisation, retaining the shape information and discarding the texture can potentially impact performance in the simulation; however, this also allows shape features to be extracted from the real domain so that $\mathcal{F}$ can operate invariantly on both domains.

\subsection{Symmetrical Adversarial Architecture}
\label{ssec:dual_adversarial_architecture}
The mapping function $\mathcal{F}$ is a pretrained EMP-Net~\cite{avraham2019empnet} using the loss function in Eq.~\ref{eq:emp_loss}, trained on RGB-D sequences from the source domain. In this work, we slightly diverge from the original method outlined in~\cite{avraham2019empnet}. The input to the EMP-Net CNN is an RGB image and not RGB-D. The depth is still required for applying the loss defined in Eq.~\ref{eq:emp_loss} on the CNN but it is not used for inferring representations from a given RGB image. This choice of input aligns well with the available information contained in the target domain.

Provided images from the source domain, $\{I_{s,i}\}_{i=0}^{LN_{s}}$, and target domain, $\{I_{t,i}\}_{i=0}^{N_{t}}$, we define a dual adversarial optimisation scheme which minimises each domain representation $p(\bar{\ell}|\mathcal{I}_{s})$ and $p(\bar{\ell}|\mathcal{I}_{t})$ towards the shared domain representation. As seen in Fig.~\ref{fig:system_overview}, there are three main components in our system: the EMP-Net CNN, which defines the initial mapping function $\mathcal{F}$ and operates as a generator in a GAN setup and two discriminators $D_{s\rightarrow t}$ and $D_{t \rightarrow s}$, where each discriminator is responsible for pulling the generator towards the shared representation distribution $\bar{\ell} \sim p(\bar{\ell}|\mathcal{I}_{s}) = p(\bar{\ell}|\mathcal{I}_{t})$.
The loss applied to $D_{t \rightarrow s}$, between the target and source domains is defined as:
\begin{equation}
\begin{split}
V_{D_{t \rightarrow s}} = \frac{1}{\mathcal{B}}\sum_{i=1}^{\mathcal{B}}[\alpha \log D_{t \rightarrow s}(\mathcal{F}(I_{s,i})) -D_{t \rightarrow s}(\mathcal{F}(I_{t,i}))]
\label{eq:source_to_shared_optim}
\end{split}
\end{equation}
Similarly, the loss applied to $D_{s \rightarrow t}$, between the source and target domain is defined as:
\begin{equation}
\begin{split}
V_{D_{s \rightarrow t}} = \frac{1}{\mathcal{B}}\sum_{i=1}^{\mathcal{B}}[\beta \log D_{s \rightarrow t}(\mathcal{F}(I_{t,i})) - D_{s \rightarrow t}(\mathcal{F}(I_{s,i}))]
\label{eq:target_to_shared_optim}
\end{split}
\end{equation}
The adversarial loss on the mapping function $\mathcal{F}$ is the loss we would apply to a generator in a dual discriminator GAN setup:
\begin{equation}
\begin{split}
GEN_{loss} = \frac{1}{\mathcal{B}}\sum_{i=1}^{\mathcal{B}}[\alpha \log D_{t \rightarrow s}(\mathcal{F}(I_{s,i})) \\ 
+ \beta \log D_{s \rightarrow t}(\mathcal{F}(I_{t,i})) \\ 
- D_{t \rightarrow s}(\mathcal{F}(I_{t,i})) \\ 
- D_{s \rightarrow t}(\mathcal{F}(I_{s,i}))]
\label{eq:dual_loss}
\end{split}
\end{equation}
where $\mathcal{B}$ is the batch size. The final loss applied to the mapping function $\mathcal{F}$ is the adversarial loss from Eq.~\ref{eq:dual_loss}, regularised by the original cross-entropy loss from Eq.~\ref{eq:emp_loss}:
\begin{equation}
\begin{split}
V_{\mathcal{F}} = GEN_{loss} + \lambda_{CE}CE_{loss}
\label{eq:uda_loss}
\end{split}
\end{equation}
where $\lambda_{CE}$ is a regularisation term defined in Section~\ref{sec:experiments}.

\section{Experiments}
\label{sec:experiments}
We perform extensive experiments for evaluating our method. This includes assessing the attributes of the adapted representations, and performing localisation on representations that have been optimised for a simulation environment and adapted to a real environment. Finally, we ablate various adaptation methods which show the effectiveness of our symmetrical adversarial approach.

\subsection{Datasets}
\label{ssec:datasets}
The simulation/source environments which were chosen for this work both use the Habitat~\cite{savva2019habitat} simulation environment. 
Habitat provides access to RGB and Depth sensors which are perfectly aligned, along with odometry. The size of the RGB and Depth images can be specified in the simulation environment, which are both set to be $256 \times 256$. 

\begin{figure}[h]
\centering
\subfloat[Matterport3D]{%
\includegraphics[width=0.22\textwidth]{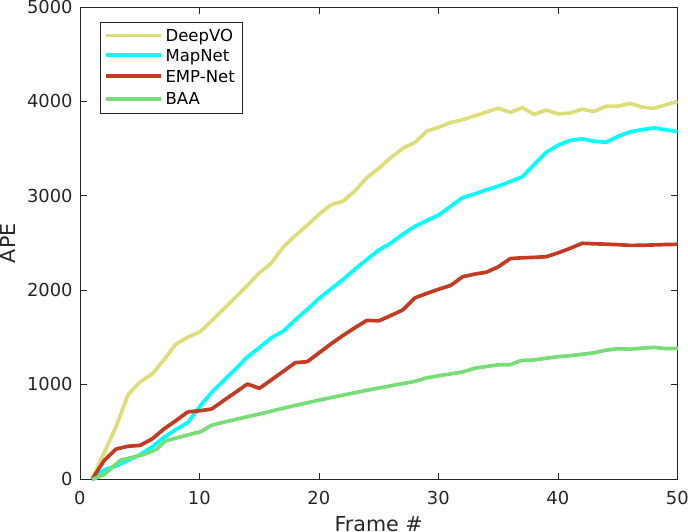}
\label{fig:mp3d_domain_generalisation}  
}
\hspace{1.0em}
\subfloat[Replica]{%
\includegraphics[width=0.22\textwidth]{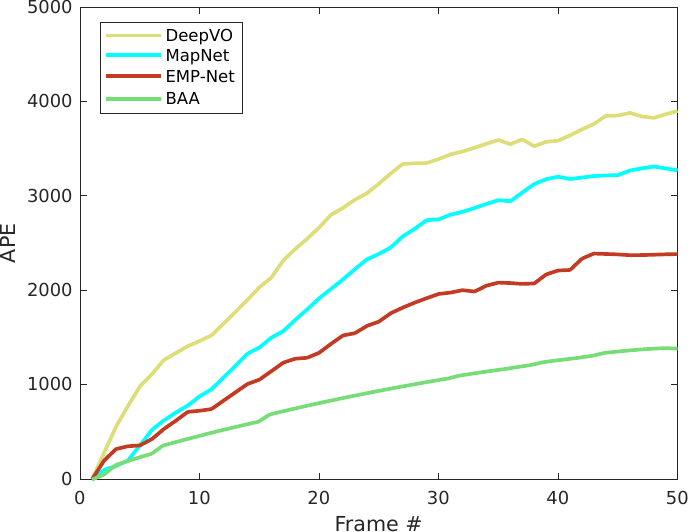}
\label{fig:replica_domain_generalisation}
}
\caption{Performance of BAA and baseline models as indicated by Average Positional Error (APE) over different sequence lengths (up to 50 frames) on the Active Vision dataset when trained on (a) Matterport3D and (b) Replica datasets. Our BAA approach offers a significant improvement over the compared baselines when no label information from the target dataset is given.}
\label{fig:domain_generalisation}
\vspace{1.0em}
\end{figure} 

\paragraph{Matterport3D}
The Matterport3D (MP3D) dataset~\cite{chang2017matterport3d} contains 90 different scenes with varying navigable space. There are 58 scenes in the training split which contain 4.8M trajectories of variable length (normally above 5 frames) which is publicly available\footnote{https://github.com/facebookresearch/habitat-api} and is part of the point-goal navigation task~\cite{savva2019habitat}. From those 4.8M trajectories, we randomly sampled 50k uniformly distributed trajectories across the scenes and randomly chose slices of 5 consecutive frames from within those 50k trajectories. This results in 200k trajectories of length 5 used for training. For the test set, we use the test split which consists of 18 scenes and 1008 combined trajectories for all the scenes. From those, we sample 3 trajectories per test scene which has a trajectory length of at least 50 frames.

\begin{table*}[h]
\begin{center}
\setlength\tabcolsep{0.3cm}
\scalebox{0.85}{
\begin{tabular}{c c c c c c c}
\specialrule{.2em}{.1em}{.1em}
& \multicolumn{3}{c}{\textit{Matterport3D}~\cite{chang2017matterport3d}} & \multicolumn{3}{c}{\textit{Replica}~\cite{straub2019replica}} \\
\cmidrule(l){2-4} \cmidrule(l){5-7}
Model &\thead{APE-5} &\thead{APE-50} &\thead{ATE-50} & \thead{APE-5} &\thead{APE-50} &\thead{ATE-50} \\
\hline
DeepVO~\cite{wang2017deepvo} & 1106.3 & 4011.5 & 1348.8 & 904.1 & 3861.2 & 1289.9 \\
MapNet~\cite{henriques2018mapnet} & 208.1 & 3669.1 & 1050.2 & 310.4 & 3288.0 & 933.3 \\
EMP-Net~\cite{avraham2019empnet} & 335.8 & 2509.7 & 841.3 & 304.1 & 2336.9 & 801.8 \\
BAA (Ours) & \textbf{245.1} & \textbf{1387.2} & \textbf{475.4} & \textbf{221.6} & \textbf{1345.8} & \textbf{470.1} \\
\hline
\end{tabular}
}
\vspace{1.0em}
\caption{Domain Generalisation when training on the Matterport3D and Replica source datasets and evaluated on the Active Vision target dataset using Average Position Error (APE) and Absolute Trajectory Error (ATE).} 
\label{tab:domain_generalisation_avd}
\end{center}
\end{table*}

\begin{figure*}[h]
\centering
\hspace{-1.0em}
\subfloat[BAA vs. Supervised Baselines]{
\begin{tabular}{c}
\includegraphics[width=0.27\textwidth]{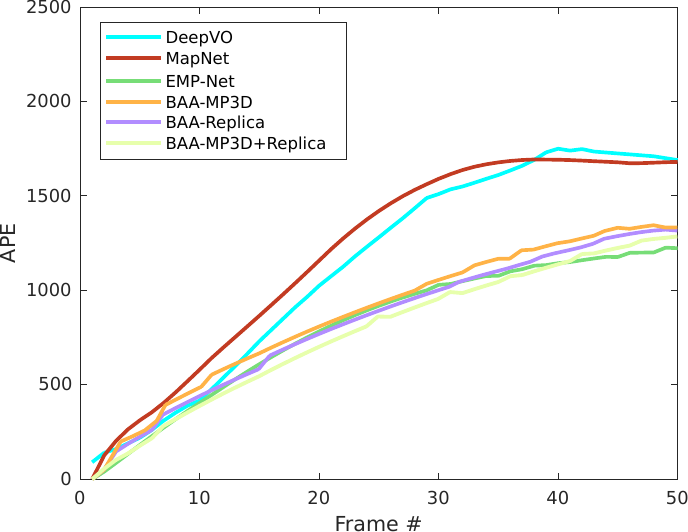}
\end{tabular}
\label{fig:supervised_ape}
}
\subfloat[Matterport3D Ablation]{
\begin{tabular}{c}
\includegraphics[width=0.27\textwidth]{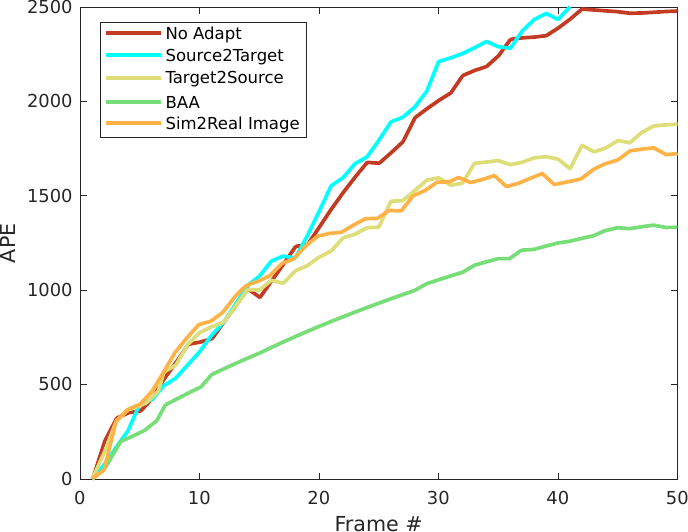}
\end{tabular}
\label{fig:avd_adaptation_ablation_mp3d}
}
\subfloat[Replica Ablation]{
\begin{tabular}{c}
\includegraphics[width=0.27\textwidth]{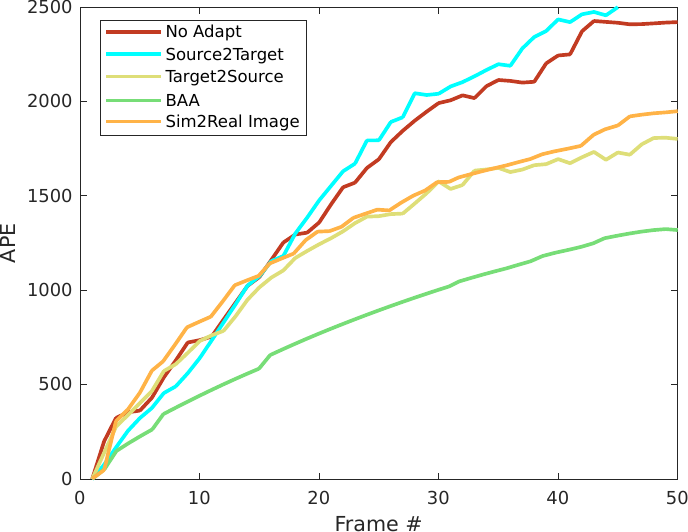}
\end{tabular}
\label{fig:avd_adaptation_ablation_replica}
}
\caption{(a) Performance of BAA and baseline models as indicated by Average Positional Error (APE) over different sequence lengths (up to 50 frames) on the Active Vision dataset. Each competing baseline model was trained under supervision using labels from the target domain. Note the performance of our BAA approach which is able to outperform all supervised baseline models except the model in~\cite{avraham2019empnet}, without \textit{any label information} from the target domain. (b) Ablation performance across different adaptation schemes as indicated by Average Positional Error (APE) over different sequence lengths (up to 50 frames) on the Active Vision dataset when trained on Matterport3D (\textbf{left}) and Replica (\textbf{right}) datasets. Our BAA approach significantly outperforms other adaptation approaches, including the image adaptation approach of~\cite{huang2018multimodal}.}
\end{figure*} 

\paragraph{Replica}
The Replica dataset~\cite{straub2019replica} contains 12 different scenes with an additional 6 scenes (``FRL apartment") where the same scene is used, but the layout is changed. This totals to 18 scenes with a smaller amount of navigable space per scene on average when compared to MP3D~\cite{chang2017matterport3d}. For this dataset there were no publicly available point-goal task navigation trajectories; as part of this work, trajectories were created. We took 11 out of the 12 scenes (excluding ``office 1" due to size) and randomly generated 50k trajectories of variable length, uniformly distributed among the scenes. A single trajectory was generated by randomly selecting two navigable points, computing the shortest path and recording the agent's traversal. Following that, similar to the procedure that was performed for MP3D, we randomly chose slices of 5 consecutive frames and construct 200k trajectories of length 5. For the test set, we use the 6 scenes of the ``FRL apartment" and follow a similar procedure of selecting two random navigable points and recording the agent's traversal, but impose a minimal length of 50 frames on a sequence. As result, 50 test trajectories are recorded with a length of 50 frames per trajectory.

\paragraph{Active Vision}
The Active Vision Dataset (AVD)~\cite{ammirato2017dataset} is chosen as the real/target domain and consists of 20 indoor scenes. We use the same test (``Home 001 1" and ``Home 001 2") and train (the rest of scenes) splits as in~\cite{henriques2018mapnet,avraham2019empnet} except that for training we do not generate trajectories. Instead, we use the entire set of images from each scene as images which represent the target domain. This amounts to a little less than 23k images. For the test set, we use the two test scenes and generate 50 random trajectories of 50 frames each. The RGB images are downsampled from $1920 \times 1080$ by a factor of $4$ and then center cropped to obtain images of $256 \times 256$. The depth information is not used, as it is only needed for the source domain.

\begin{table*}[h]
\begin{center}
\setlength\tabcolsep{0.2cm}
\scalebox{0.85}{
\begin{tabular}{c c c c c c c}
\specialrule{.2em}{.1em}{.1em}
& \multicolumn{3}{c}{\textit{Matterport3D}~\cite{chang2017matterport3d}} & \multicolumn{3}{c}{\textit{Replica}~\cite{straub2019replica}} \\
\cmidrule(l){2-4} \cmidrule(l){5-7}
Model &\thead{APE-5} &\thead{APE-50} &\thead{ATE-50} & \thead{APE-5} &\thead{APE-50} &\thead{ATE-50} \\
\hline
EMP-Net~\cite{avraham2019empnet} (No Adaptation) & 335.8 & 2509.7 & 841.3 & 304.1 & 2336.9 & 801.8 \\
EMP-Net~\cite{avraham2019empnet} (Sim2Real~\cite{huang2018multimodal}) & 375.2 & 1707.3 & 699.7 & 437.2 & 1967.6 & 789.2 \\
EMP-Net (S2T) & 375.1 & 2656.2 & 877.7 & 314.2 & 2669.1 & 886.9 \\
EMP-Net (T2S) & 361.3 & 1925.3 & 752.3 & 402.1 & 1819.6 & 724.2 \\
BAA (Ours) & \textbf{245.1} & \textbf{1387.2} & \textbf{475.4} & \textbf{221.6} & \textbf{1345.8} & \textbf{470.1} \\
\hline
\end{tabular}
}
\vspace{1.0em}
\caption{Ablation of different adaptation schemes on the EMP-Net baseline trained on (a) Matterport3D and (b) Replica datasets and evaluated using Average Position Error (APE) and Absolute Trajectory Error (ATE). Our symmetrical adaptation approach significantly outperforms other adaptation approaches, including the image adaptation approach of~\cite{huang2018multimodal}.} 
\label{tab:avd_adaptation_ablation}
\end{center}
\end{table*}

\subsection{Implementation}
\label{ssec:imp_details}

\paragraph{Localisation Network Architecture}
To extract feature embeddings, we employ the same CNN architecture as~\cite{avraham2019empnet} which is based off the U-Net architecture~\cite{ronneberger2015u}, initialised using the approach in~\cite{he2015delving}. The U-Net architecture consists of an encoder and decoder network where the encoder network contains U-Net blocks and U-Net downsampling blocks. Each U-Net block consists of a repeated sequence of a $3\times3$ convolution layer, followed by BatchNorm~\cite{ioffe2015batch} and ReLU activation. For each U-Net downsampling block, the U-Net block is used where the first $3\times3$ convolution layer in the block is modified to use 2-strided convolutions to downsample the input. The number of output channels is upsampled with each U-Net downsampling block by a factor of 2 starting with a base output channel of $8$ and upsampled to $128$ output channels. For the decoder, we used a U-Net upsampling block, which is a sequence consisting of a $2\times2$ transposed convolution with stride 2 followed by a U-Net block. The output of each U-Net upsampling block is then concatenated to its matching output from the encoder network. The localisation network also functions as the generator in our GAN setup where the choice of loss is consistent with the objective function defined in Eq.~\ref{eq:uda_loss}.

\paragraph{GAN architecture}
Our GAN setup consists of a set of dual discriminator networks paired with a generator network (which also serves as our representation extractor). Each discriminator network uses the architecture of~\cite{radford2015unsupervised}, which consists of 2-strided $5\times5$ convolution layers with Leaky ReLU activation to downsample our input RGB images from $256\times256$ to $4\times4$ in height and width. 
Following the setup of~\cite{nguyen2017dual}, we use Batch Normalisation~\cite{ioffe2015batch} after each convolutional layer in both discriminators (with the exception of the first layer). 
Our choice of adversarial loss is consistent with the objective functions defined in Eqs.~\ref{eq:source_to_shared_optim} \&~\ref{eq:target_to_shared_optim}.

\paragraph{Training Settings}
For training the localisation network, we follow the settings in~\cite{avraham2019empnet}, training with batch size of 16, where each sample in the batch in a sequence of 5 consecutive RGB frames. RGB and depth data were normalised to lie between $[0,1]$. We maintain the buffer size $b$ of~\cite{avraham2019empnet} to 4 frames, and extract 1024 feature representations for each observation frame. We use the ADAM optimiser~\cite{kingma2014adam}, with first and second momentum terms of $0.5$ and $0.999$ values respectively. Across all experiments, we choose $\lambda_{CE} = 0.1$, which was found to be the optimal value via performing a random walk parameter sweep of $\lambda_{CE}$ values in a range over $[0.001,10]$. We set $\alpha=0.02$ and $\beta=0.04$; similarly these values were found through a random walk parameter sweep of $\alpha$ and $\beta$ values in a range over $[0.001,0.5]$. These values of $\alpha$ and $\beta$ are consistent with what we would expect; since there is a regularising term $L_{CE}$ (Eq.~\ref{eq:emp_loss}) which affects $p(\bar{\ell}|\mathcal{I}_s)$ from the source domain, in order to \textit{balance} the adaptation, a larger $\beta$ (attracting force towards the target domain) was found to be more effective. 

We first pretrain our localisation network with a learning rate of $10^{-3}$ for $10$ epochs on source domain data before adaptation. For adaptation, we train with a batch size of 32. We employ the Two-Time Update Rule (TTUR)~\cite{heusel2017gans}, using a 3:1 ratio in learning rates between the discriminator and generator networks of $3e\text{-}4$ and $1e\text{-}4$ respectively. In this GAN setup, our pretrained localisation network acts as the generator network which is concurrently trained with both discriminator networks for 10 epochs on the target domain data.

\subsection{Adapting Properties of Representations}
\label{ssec:representation_distribution}

\begin{figure}
\centering
\subfloat[]{%
  \includegraphics[width=0.22\textwidth]{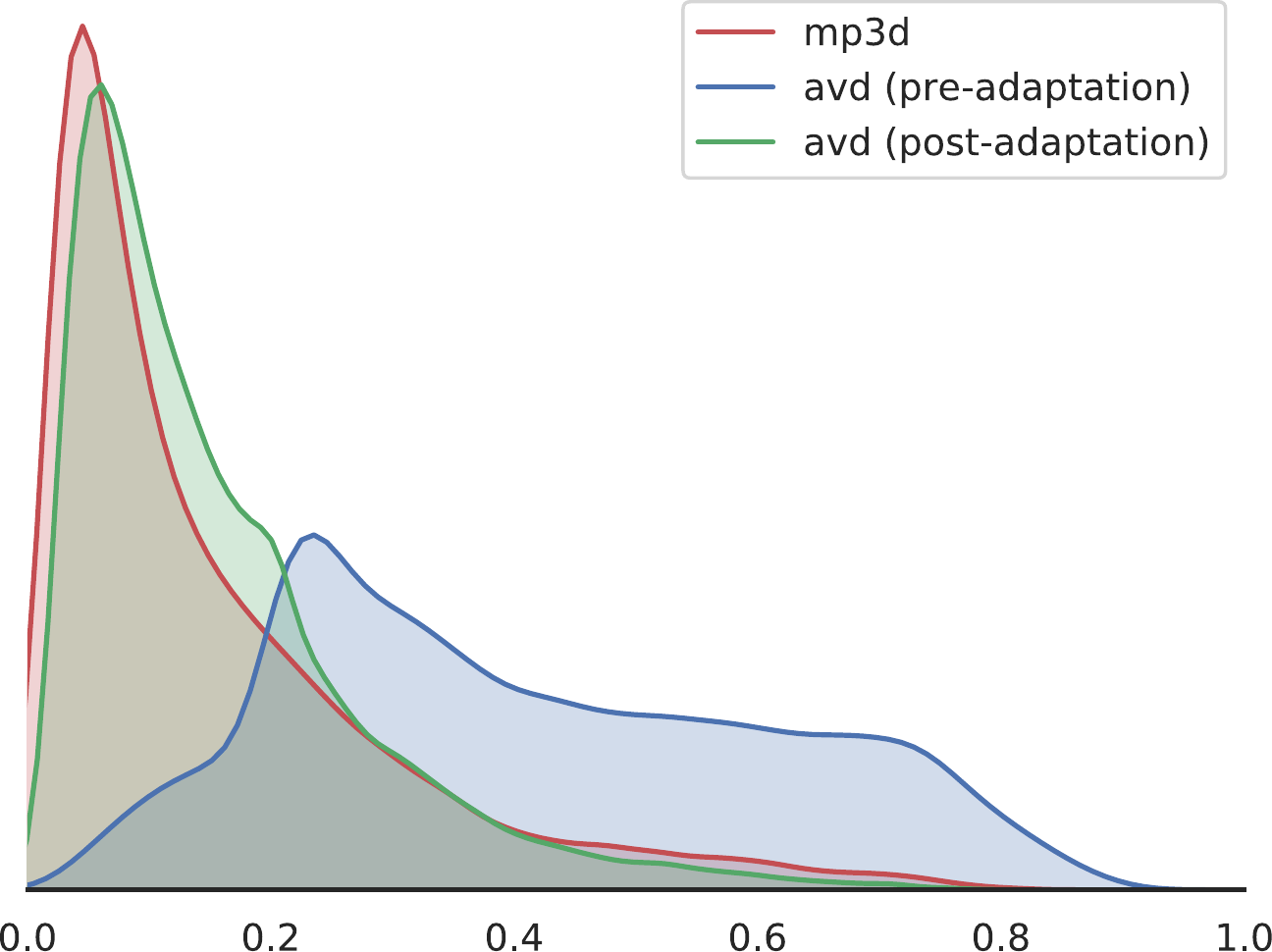}
  \label{fig:mp3d_KDE_plots}
  }
\subfloat[]{%
  \includegraphics[width=0.22\textwidth]{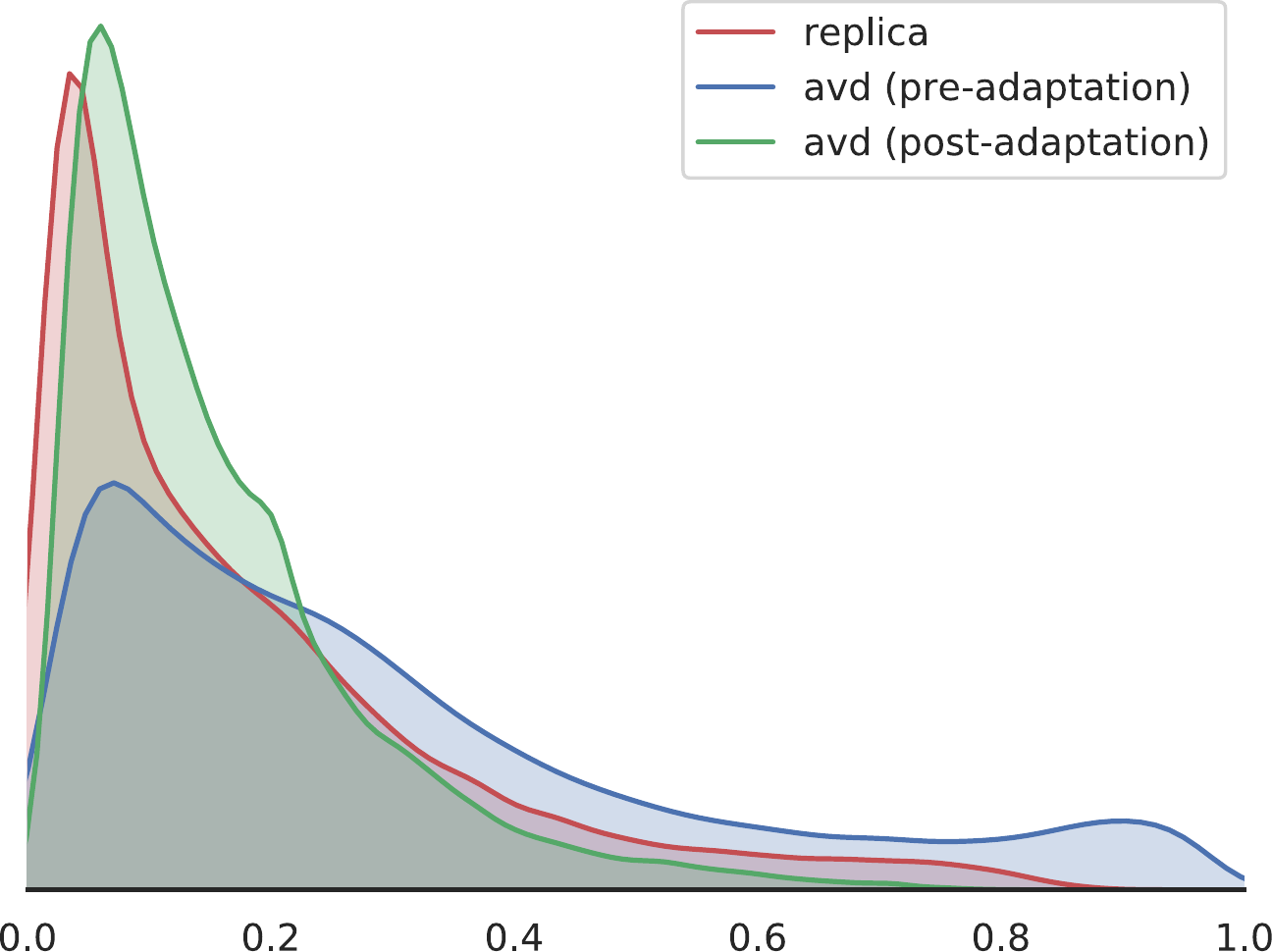}
 \label{fig:replica_KDE_plots}
 }
\caption{Histogram plots of the mutual exclusiveness property between representations. This property is important for generating unique descriptors which are effective for finding correspondences. The histogram of a pretrained EMP-Net~\cite{avraham2019empnet} on two simulation environments (MP3D and Replica) can be seen in (a) and (b) respectively, along with pre-adaptation and post-adaptation histograms to AVD. 
}
\label{fig:KDE_plots}
\vspace{1.0em}
\end{figure} 

An important property that the adapted representations need to maintain is uniqueness; or as mentioned in Section~\ref{sec:background}, in a given image every two representation vectors should be mutually exclusive. In this experiment, we test whether this property is preserved after performing the adaptation on a target domain. 

Generally, for a single image, this is done by taking the negative pair-wise distance between the representations, performing a Softmax operation and measuring the average value for each representation \textit{w.r.t} every other representation. Ideally, this value should be $0$. This process can be repeated for multiple images, which allows us to accumulate and produce a histogram of the aforementioned property. 
Four experiments are shown in Figs.~\ref{fig:mp3d_KDE_plots} \&~\ref{fig:replica_KDE_plots}, where the source domains are MP3D and Replica respectively. The histogram for both source domains is shown, and we observe that the mutual exclusiveness property is preserved, similar to~\cite{avraham2019empnet}. Furthermore, the histograms for the target domain, AVD, are shown for MP3D (Fig.~\ref{fig:mp3d_KDE_plots}) and Replica (Fig.~\ref{fig:replica_KDE_plots}) before performing our method (AVD pre-adaptation) and after (AVD post-adaptation). The effectiveness of our approach can be seen as the mutual exclusiveness property between representations is more concentrated near $0$ post-adaptation in relation to pre-adaptation where it is significantly weaker.

\subsection{Evaluation on Real-World Data}
\label{ssec:sim2real_results}
\paragraph{Training on Simulation and Evaluating on Real}
In Fig.~\ref{fig:domain_generalisation} and Tab.~\ref{tab:domain_generalisation_avd}, we compare against the methods from~\cite{wang2017deepvo},~\cite{henriques2018mapnet} and~\cite{avraham2019empnet} when trained on source domain simulation data from Matterport3D and Replica and evaluated on target domain real-world data from AVD. We can observe that without label information from the target domain, our BAA method substantially outperforms competing baselines.

\paragraph{Evaluating against Fully-Supervised methods}
In Fig.~\ref{fig:supervised_ape} and Table~\ref{tab:avd_ape_ate}, we compare our BAA method against the methods of~\cite{wang2017deepvo},~\cite{henriques2018mapnet} and~\cite{avraham2019empnet} when each competing method is trained in a fully-supervised manner on training data from the target domain. Additionally, to evaluate the performance of our method against a pure geometrical approach, we compare to~\cite{mur2015orb}, where we show that we are able to significantly outperform ORB-SLAM2 on the AVD dataset. Note that in this case, we show the \textit{unsupervised} performance of our method on the target domain (our BAA model is not given target domain labels). We can see that our BAA method compares very favourably against the fully-supervised method, outperforming the supervised methods of~\cite{wang2017deepvo} and~\cite{henriques2018mapnet} and obtaining competitive results with the supervised method from~\cite{avraham2019empnet}. This result highlights the ability to leverage noiseless depth measurements provided in the simulation environment for learning representations on real-world RGB measurements; in contrast to the fully supervised methods which rely on depth information that is noisy and comes from depth sensors. We reiterate that BAA \textit{does not use depth information} as an input to our framework in the inference stage.

\begin{table}[t]
\small
\begin{center}
\setlength\tabcolsep{0.1cm}
\scalebox{1.0}{
\begin{tabular}{c c c c c}
\specialrule{.2em}{.1em}{.1em}
\thead{AVD data~\cite{ammirato2017dataset}} &\thead{APE-5} &\thead{APE-50} &\thead{ATE-50} \\
\hline
ORB-SLAM2 (RGB-D)~\cite{mur2015orb} & 432.0  & 3090.0 & 794.0 \\
DeepVO~\cite{wang2017deepvo} & 220.0 & 1690.0 & 741.0 \\
MapNet~\cite{henriques2018mapnet} & 312.3 & 1680.0 & 601.0 \\
EMP-Net~\cite{avraham2019empnet} & \textbf{181.6} & \textbf{1201.0} & \textbf{381.0} \\
BAA-MP3D (Ours) & 245.1 & 1387.2 & 475.4 \\
BAA-Replica (Ours) & 221.6 & 1345.8 & 470.1 \\
BAA-MP3D+Replica (Ours) & 185.2 & 1286.4 & 433.8 \\
\hline
\end{tabular}
}
\vspace{1.0em}
\caption{Average Position Error (APE) and Absolute Trajectory Error (ATE) on the Active Vision Dataset. Note that all competing baselines have been trained using label information from the Active Vision Dataset, whilst our BAA approach has only been trained using source domain data (Matterport3D and Replica datasets) and adapted for localisation on the Active Vision Dataset.}
\label{tab:avd_ape_ate}
\vspace{1.0em}
\end{center}
\end{table}

\subsection{Ablation of Different Adaptation Schemes}
\label{ssec:adaptation_ablations}
In Figs.~\ref{fig:avd_adaptation_ablation_mp3d} \&~\ref{fig:avd_adaptation_ablation_replica} and Table~\ref{tab:avd_adaptation_ablation}, we show ablation results for different schemes adapting from Matterport3D and Replica to AVD. We observe that our BAA approach offers significant improvements in terms of APE and ATE performance than uni-directional adaptation approaches~\cite{shen2017adversarial} and an image adaptation approach~\cite{huang2018multimodal}. This experiment highlights the  key intuition, that if two environments share geometrical similarities and differ in textural and lighting appearances, using a symmetrical adversarial loss encourages the generator to perform a decoupling of these two properties.


\section{Conclusion}
\label{sec:conclusion}
In this paper, we presented a domain adaptation approach which uses a symmetrical distance between source and target domains for adapting representations between domains. We show that by adapting directly on localisation features optimised on simulation data and using our symmetrical adaptation approach, we can gain large improvements in localisation performance on real-world data without any label information from the target domain. We show that our approach is not only effective in an unsupervised learning setting, but can achieve very favourable results when compared to supervised methods on real-world data.

\paragraph{Acknowledgments}
The authors would like to thank the anonymous reviewers for their useful and constructive comments. This work was supported by the Australian Research Council Centre of Excellence for Robotic Vision (project number CE1401000016).

\clearpage
%
%
\bibliographystyle{ieee}
\bibliography{egbib}
\end{document}